\documentclass[runningheads]{llncs}
\PassOptionsToPackage{dvipsnames,table}{xcolor}
\usepackage{graphicx}

\usepackage{tikz}
\usepackage{comment}
\usepackage{amsmath,amssymb} 
\usepackage{color}
\usepackage{savesym}
\savesymbol{checkmark}
\usepackage{dingbat}
\usepackage{array}
\usepackage{tabularx}
\usepackage{subfigure}
\usepackage{caption}
\usepackage{graphicx}
\usepackage{booktabs}
\usepackage{multirow}
\usepackage{siunitx}
\usepackage[square, numbers]{natbib}
\usepackage{floatrow}
\usepackage{blindtext}
\usepackage[accsupp]{axessibility}  

\usepackage[dvipsnames,table]{xcolor}
\usepackage[pagebackref]{hyperref}
\hypersetup{
    colorlinks=true,
    citecolor=ForestGreen,
    linkcolor=blue,
    filecolor=magenta,      
    urlcolor=cyan
}

\usepackage{gensymb}

\begin{document}
\pagestyle{headings}
\mainmatter
\def\ECCVSubNumber{18}  

\title{UAV-based Visual Remote Sensing for Automated Building Inspection} 

\titlerunning{UVRSAB}
%
\author{Kushagra Srivastava\inst{1}\thanks{denotes equal contribution} \index{Srivastava, Kushagra}\and
Dhruv Patel \index{Patel, Dhruv}\inst{1}\textsuperscript{$\star$} \and
Aditya Kumar Jha\index{Jha, Aditya Kumar}\inst{2} \and
Mohhit Kumar Jha \index{Jha, Mohhit Kumar} \inst{2}\and
Jaskirat Singh \index{Singh, Jaskirat}\inst{3} \and
Ravi Kiran Sarvadevabhatla \index{Sarvadevabhatla, Ravi Kiran}\inst{1} \and
Pradeep Kumar Ramancharla \index{Ramancharla, Pradeep Kumar}\inst{1}
\and 
Harikumar Kandath \index{Kandath, Harikumar}\inst{1}
\and
K. Madhava Krishna \index{Krishna, K. Madhava}\inst{1}
} 
\authorrunning{K. Srivastava et al.}
%
\institute{International Institute of Information Technology, Hyderabad, India  
\email{\{kushagra2000, dhruv.r.patel14\}@gmail.com} \\
\email{\{ravi.kiran, harikumar.k, ramancharla, mkrishna\}@iiit.ac.in}\\\and
Indian Institute of Technology, Kharagpur, India \\
\email{\{aditya.jha, mohhit.kumar.jha2002\}@kgpian.iitkgp.ac.in}\and
University of Petroleum and Energy Studies, Dehradun, India\\\email{juskirat2000@gmail.com}}
\maketitle

\begin{abstract}
Unmanned Aerial Vehicle (UAV)  based remote sensing system incorporated with computer vision has demonstrated potential for assisting building construction and in disaster management like damage assessment during earthquakes. The vulnerability of a building to earthquake can be assessed through inspection that takes into account the expected damage progression of the associated component and the component’s contribution to structural system performance. Most of these inspections are done manually, leading to high utilization of manpower, time, and cost. This paper proposes a methodology to automate these inspections through UAV-based image data collection and  a software library for post-processing that helps in estimating the seismic structural parameters. The key parameters considered here are the distances between adjacent buildings, building plan-shape, building plan area, objects on the rooftop and rooftop layout. The accuracy of the proposed methodology in estimating the above-mentioned parameters is verified through field measurements taken using a distance measuring sensor and also from the data obtained through Google Earth. Additional details and code can be accessed from \url{https://uvrsabi.github.io/}

\end{abstract}

\section{Introduction}
Traditional techniques to analyze and assess the condition and geometric aspects of buildings and other civil structures involve physical inspection by civil experts according to pre-defined procedures. Such inspections can be costly, risky, time-consuming, labour and resource intensive. A considerable amount of research has been dedicated to automating and improving civil inspection and monitoring through computer vision. This results in less human intervention and lower cost while ensuring effective data collection. Unmanned Aerial Vehicles (UAVs) mounted with cameras have the potential for contactless, rapid and automated inspection and monitoring of civil structures as well as remote data acquisition.

Computer vision-aided civil inspection has two prominent areas of application: damage detection and structural component recognition  \cite{SPENCER2019199}. Studies focused on damage detection have used heuristic feature extraction methods to detect concrete cracks \cite{concrete1, concrete2, concrete3}, concrete spalling \cite{spalling1, spalling2}, fatigue cracks \cite{fatigue1, fatigue2} and corrosion in steel \cite{corrosion1,corrosion2,corrosion3}. However, heuristic-based methods do not account for the information that is available in regions around the defect and have been replaced with deep learning-based methods. Image classification \cite{classification1, classification2, classification3}, object detection \cite{object1, object2}, semantic segmentation \cite{semantic1, semantic2, semantic3} based methods have been used to successfully detect and classify the damage type. On the contrary, structural component analysis involves detecting, classifying and studying the characteristics of a physical structure. Hand-crafted filters \cite{filter1, filter2}, point cloud-based \cite{cloud1, cloud2, cloud3, cloud4}, and deep learning-based \cite{deep1,deep2, deep3, deep4} methods have been used to assess structural components like columns, planar walls, floor, bridges, beams and slabs. There also has been a emphasis on developing architectures for Building Information Modelling (BIM) \cite{bim1, bim2, bim3} that involves analysis of physical features of a building using high resolution 3D reconstruction.  

Apart from structural component recognition, it is also essential to assess the risk posed by earthquakes to buildings and other structural components. This is a crucial aspect of inspection in seismically active zones. Accurate seismic risk modeling requires knowledge of key structural characteristics of buildings. Learning-based models in conjunction with street imagery \cite{street1, street2} have been used to perform building risk assessments. However, UAVs can also be used to obtain information in areas difficult to access by taking a large number of images and videos from several points and different angles of view. Thus, UAVs demonstrate huge potential when it comes to remote data acquisition for pre- and/or post-earthquake risk assessments \cite{uavassessment}. 
The main contributions of this paper are given below.
\begin{enumerate}
    \item Primarily, we automate the inspection of buildings through UAV-based image data collection and a post-processing module to infer and quantify the details. This in effect avoids manual inspection, reducing the time and cost.
    \item We estimate the distance between adjacent buildings and structures. To the best of our knowledge, there has not been any work that has addressed this problem.
    \item We develop an architecture that can be used to segment roof tops in case of both orthogonal and non-orthogonal view using a state-of-the-art semantic segmentation model. 
    \item The software library for post-processing collates different algorithms used in computer vision along with UAV state information to yield an accurate estimation of the distances between adjacent buildings, building plan-shape, building plan area, objects on the rooftop, and rooftop layout. These parameters  are key for the preparation of safety index assessment for buildings against earthquakes.
\end{enumerate}
\section{Related Works}

\subsection{Distance between Adjacent Structures}
The collision between adjacent buildings or among  parts of the same building during strong earthquake vibrations is called pounding \cite{MIARI2019135}. Pounding occurs due to insufficient  physical separation  between adjacent structures and their out-of-phase vibrations resulting in non-synchronized vibration amplitudes. Pounding can lead to the generation of a high-impact force that may cause either architectural or structural damage.  Some reported cases of pounding include i) The earthquake of 1985 in Mexico City  \cite{mexico} that left more than 20\% of buildings damaged, ii) Loma Prieta earthquake of 1989 \cite{loma} that affected over 200 structures, iii) Chi-Chi earthquake of 1999 \cite{chichi} in central Taiwan, and iv) Sikkim earthquake (2006) \cite{sikkim}. Methods such as Rapid Visual Screening (RVS), seismic risk indexes, and vulnerability assessments have been developed to analyze the level of damage to a building \cite{rvs}. In particular, RVS-based methods have been used for pre-and/or post-earthquake screening of buildings in earthquake-prone areas. The pounding effect is considered as a vulnerability factor by RVS methods like FEMA P-154, FEMA 310, EMS-98 Scale, NZSEE, OSAP, NRCC, IITK-GSDMA, EMPI and RBTE-2019 \cite{primer}. 

The authors in \cite{vacca2017} present a UAV-based site survey using both Nadir and Oblique images for appropriate 3D modelling.  The integration of nadir UAV images with oblique images ensures a better inclusion of facades and footprints of the buildings. Distances between the buildings in the site were manually measured from the generated dense point cloud. We use the 3D reconstruction of the structures from images in conjunction with conditional plane fitting for estimating the distance between adjacent structures. 

\subsection{Plan Shape and Roof Area Estimation}
The relationship between the center of stiffness and gravity's eccentricity is influenced by shape irregularities, asymmetries, or concavities, as well as by building mass distributions. For any structure, if the centre of stiffness is moved away from the centre of gravity during ground motion, more torsion forces are produced \cite{arnold1982building}. When a building is shaken by seismic activity, this eccentricity causes structures to exhibit improper dynamic characteristics. Hence, the behavior of a building under seismic activity also depends on its 3D configuration, plan shape and mass distribution \cite{sahar2010using}. \textit{Plan shape and and Roof Area} is needed for calculating the Floor Space Index (FSI). FSI is the ratio of the total built-up area of all the floors to the plot area. FSI is a contributing factor in assessing the extent of the damage and is usually fixed by the expert committee.

Roof-top segmentation has been considered as a special case of 3D plane segmentation from point clouds and can be achieved through model fitting \cite{ransacplan}, region growing \cite{regionplan}, feature clustering \cite{featureplan} and global energy optimization-based methods \cite{energyplan}. Studies focused on these methods have been tested on datasets where the roof was visible orthogonally through satellite imagery\cite{satellite} and LiDAR point clouds \cite{ransacplan, regionplan, featureplan}. The accuracy of these methods depends on how the roof is viewed. In case of a non-orthogonal view, these methods must be used in conjunction with some constraints. On the contrary, learning-based methods \cite{roofdeep} have been developed that specifically segment out roofs. The neural networks employed in these methods have been trained on satellite imagery and do not perform well in non-orthogonal roof-view scenarios. Our approach is to segment out roofs when viewed both orthogonally and non-orthogonally by training a state-of-the-art semantic segmentation model on a custom roof-top dataset. 
\subsection{Roof Layout Estimation}
Roof Layout Estimation refers to identifying and locating objects present on the roof such as air conditioner units, solar panels, etc. Such objects are usually non-structural elements (NSE). As the mass of the NSE increases, the earthquake response of the NSE starts affecting the whole building. Hence, they need to be taken into account for design calculations. Furthermore, the abundance of these hazardous objects may create instabilities on the roof making it prone to damage during earthquakes. Estimating the \textit{Roof Layout} is not as trivial as in the case of satellite images, since the UAV has altitude limitations along with camera Field of View (FOV) constraints, thereby limiting us from obtaining a complete view of the roof in a single image. Moreover, we cannot rely on satellite images because it does not provide us with real time observation of our location of interest. Hence, we solve this problem by first stitching a large number of images with partially visible roofs to create a panoramic view of the roof and then we apply object detection and semantic segmentation to get the object and roof masks respectively.

Various techniques for image stitching can be roughly distinguished into three categories: direct technique \cite{6664678,2016Meas...84...32A,featurebased}, feature-based technique \cite{7838721,Lowe2004,6406008} and position-based technique \cite{tsao2018stitching}. The first category performs pixel-based image stitching by minimizing the sum of the absolute difference between overlapping pixels. These methods are scale and rotation variant and to tackle this problem, the second category focuses on extracting a set of images and matching them using feature based algorithms which includes SIFT, SURF, Harris Corner Detection. These methods are computationally expensive and fail in the absence of distinct features. The third category stitches images sampled from videos through their overlapping FOV. Due to the inability to obtain accurate camera poses, not much research has been conducted on this approach. In this paper, we present an efficient and reliable approach to make use of the camera poses and stitch a large set of images avoiding the problems of image drift and expensive computation associated with the first two categories.
\section{Data Collection}
\label{sec:datacollection}
This section discusses the methods for gathering data that were utilized to carry out the research experiments in this study.  DJI Mavic Mini\footnote{UAV specification details can be found at the official DJI website: \href{https://www.dji.com/mavic-mini}{https://www.dji.com/mavic-mini}} UAV is used for gathering visual data because of its high-quality image sensory system with an adjustable gimbal.

\begin{figure}
    \centering
    \subfigure[\textit{Frontal mode}]
    {
        \includegraphics[width=1.4in]{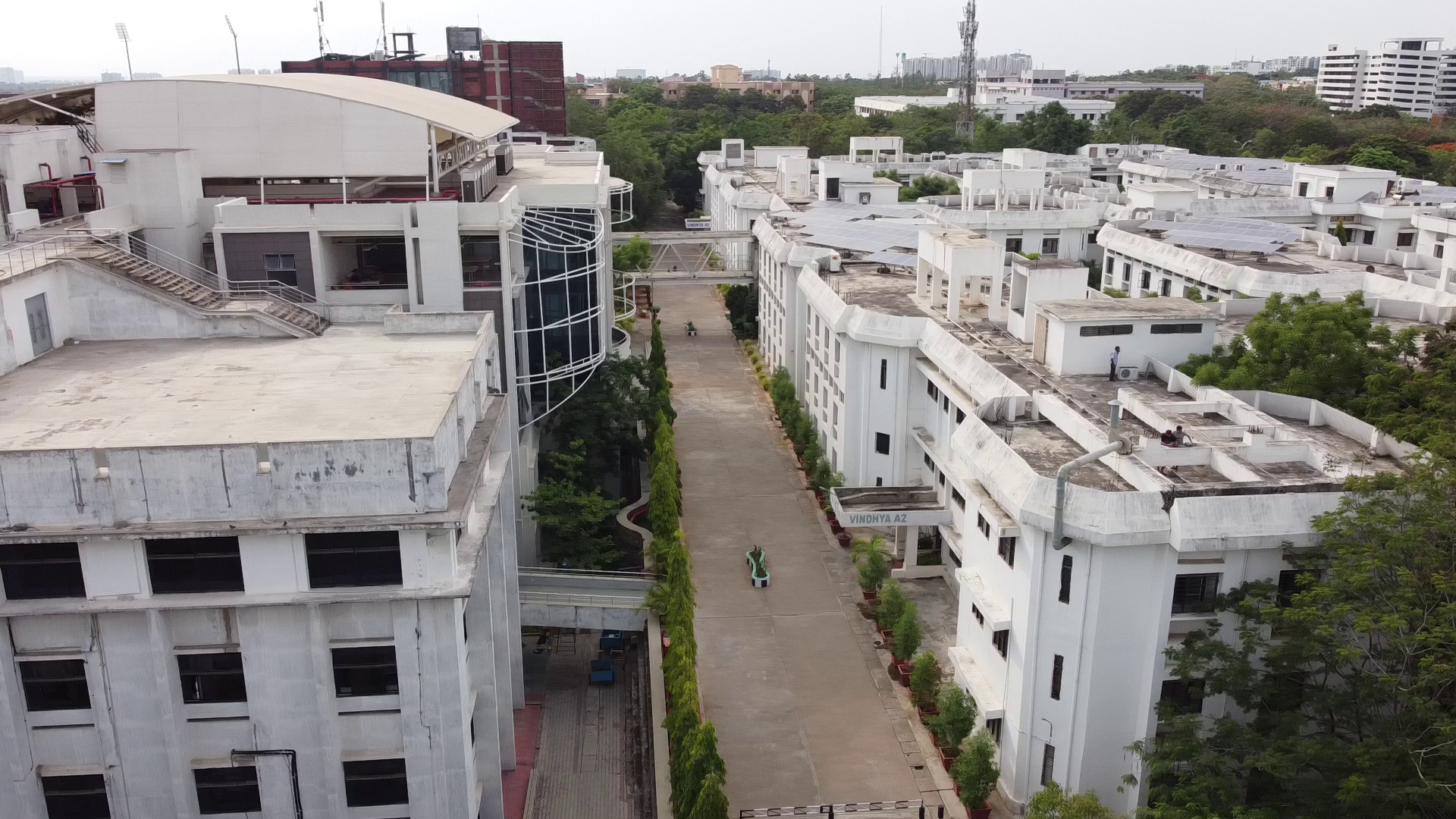}
        \label{fig:first_sub}
    }
    \subfigure[\textit{In-Between mode}]
    {
        \includegraphics[width=1.4in]{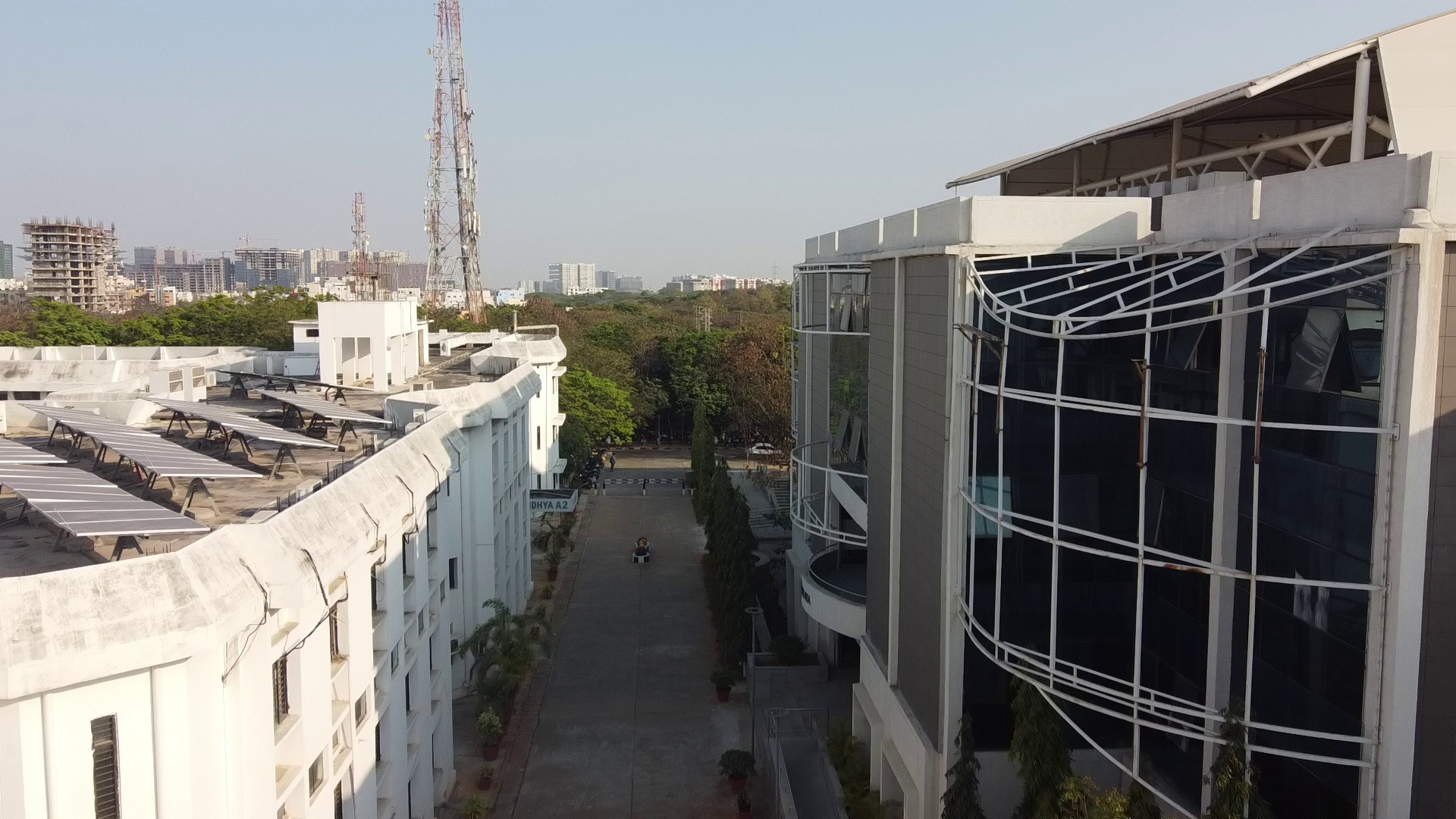}
        \label{fig:second_sub}
    }
    \subfigure[\textit{Roof mode}]
    {
        \includegraphics[width=1.4in]{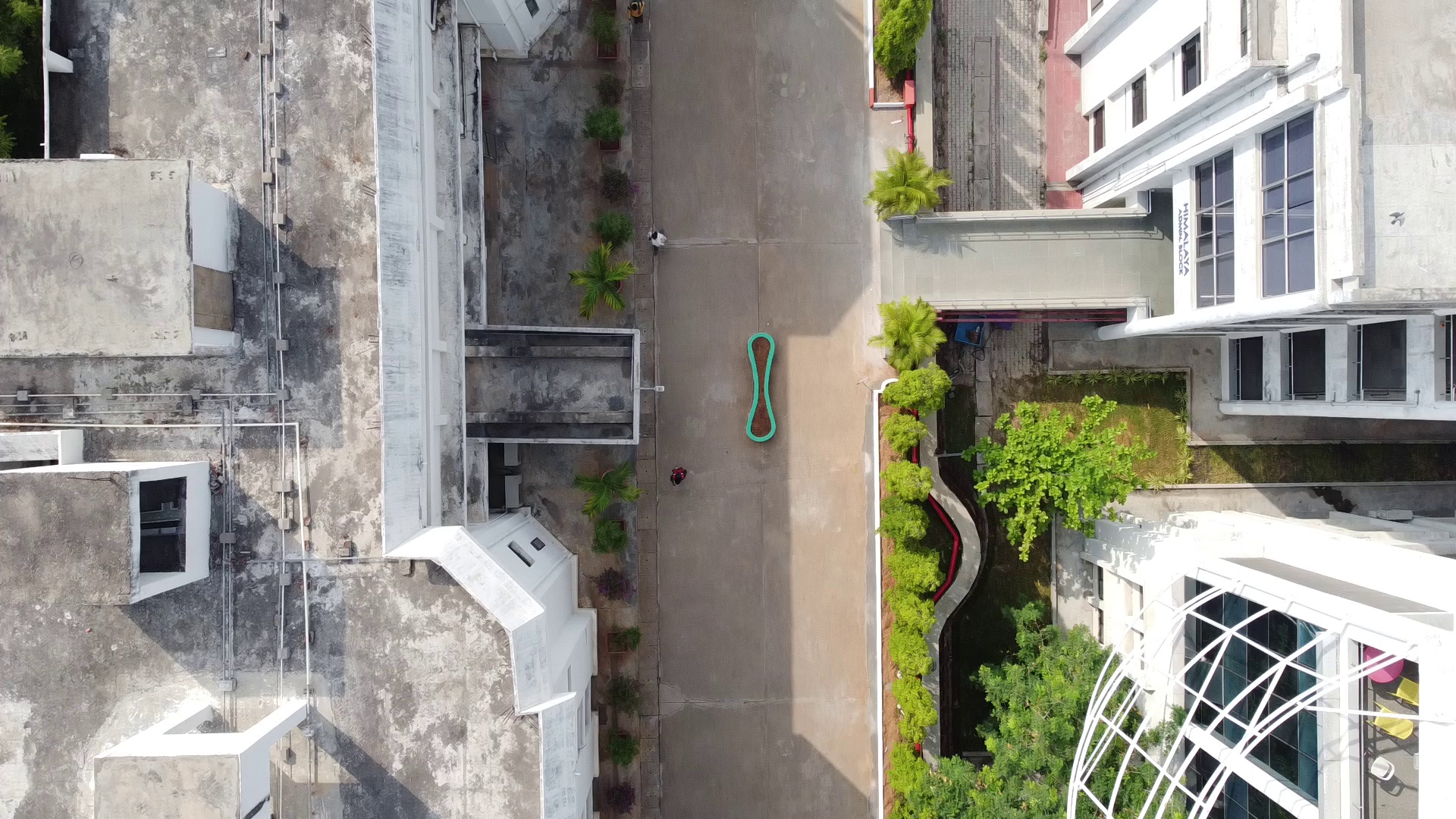}
        \label{fig:third_sub}
    }\\
        \subfigure[\textit{Frontal mode}]
    {
        \includegraphics[width=1.4in]{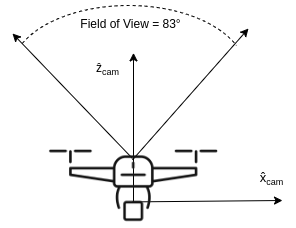}
        \label{fig:fourth_sub}
    }
    \subfigure[\textit{In-Between mode}]
    {
        \includegraphics[width=1.4in]{images/InBetweenView.png}
        \label{fig:fifth_sub}
    }
    \subfigure[\textit{Roof mode}]
    {
        \includegraphics[width=1.4in]{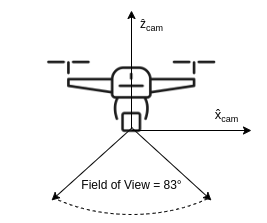}
        \label{fig:sixth_sub}
    }
    \caption{Figures \ref{fig:first_sub}, \ref{fig:second_sub}, and \ref{fig:third_sub} indicate the UAV's point of view while figures \ref{fig:fourth_sub}, \ref{fig:fifth_sub}, and \ref{fig:sixth_sub} are representations of the respective coordinate system adopted.}
    \label{distance_module}
\end{figure}
For estimating the distance between adjacent structures, the images are collected in 3 different modes: \textit{Frontal Mode}, \textit{In-Between Mode} and \textit{Roof Mode}. Fig. \ref{fig:first_sub} shows the frontal face of the two adjacent buildings for which data was collected. In this mode, we focus on estimating the distance between the two buildings by analyzing only their frontal faces through a forward-facing camera. This view is particularly helpful when there are impediments between the subject buildings and  flying a UAV between them is challenging. In fig. \ref{fig:second_sub}, the UAV was flown in-between the two buildings along a path parallel to the facade with a forward-facing camera. This mode enables the operators to calculate distances when buildings have irregular shapes. Lastly, for the \textit{roof mode}, the UAV was flown at a fixed altitude with a downward-facing camera so as to capture the rooftops of the subject buildings. Fig. \ref{fig:third_sub} is a pictorial representation of the \textit{roof mode}. The \textit{roof mode} helps in tackling occlusions due to vegetation and other physical structures. 

For \emph{Rooftop Layout Estimation}, the UAV was flown at a constant height with a downward-facing camera, parallel to the plane of the roof. This helped in robust detection of NSE. To estimate the \textit{Plan Shape and Roof Area}, a dataset comprising of around 350 images was prepared from the campus buildings and UrbanScene3D dataset \cite{UrbanScene3D}. The training set comprised of images scraped from the UrbanScene3D videos, \textit{Buildings 4} and \textit{6} and  while the validation and test set comprised of the \textit{Buildings 3}, \textit{5} and \textit{7}. This was done to ensure that the model learns the characteristic features of a roof irrespective of the building plan shape. Out of these, 50 images had fully-visible buildings while the rest contained partially-visible buildings. 

\section{Methodology}
\begin{figure}[!htb]
    \centering
    \includegraphics[height=4cm]{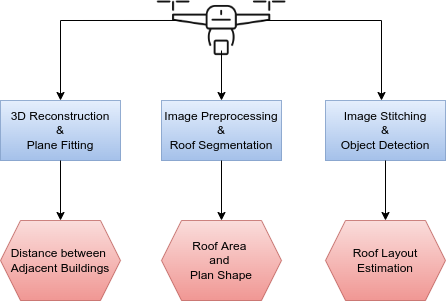}
    \caption{Architecture of automated building inspection using the aerial images captured using UAV. The odometry information of UAV is also used for the quantification of different parameters involved in the inspection.}
    \label{fig:architecture}
\end{figure}

We propose different methods to calculate the distance between the adjacent buildings using plane segmentation; estimate the roof layout using Object Detection and large scale image stitching; estimate the roof area and plan shape using roof segmentation as shown in Fig. \ref{fig:architecture}.
\subsection{Distance Between Adjacent Buildings}
\begin{figure}[!htb]
    \centering
    \includegraphics[width = 12cm]{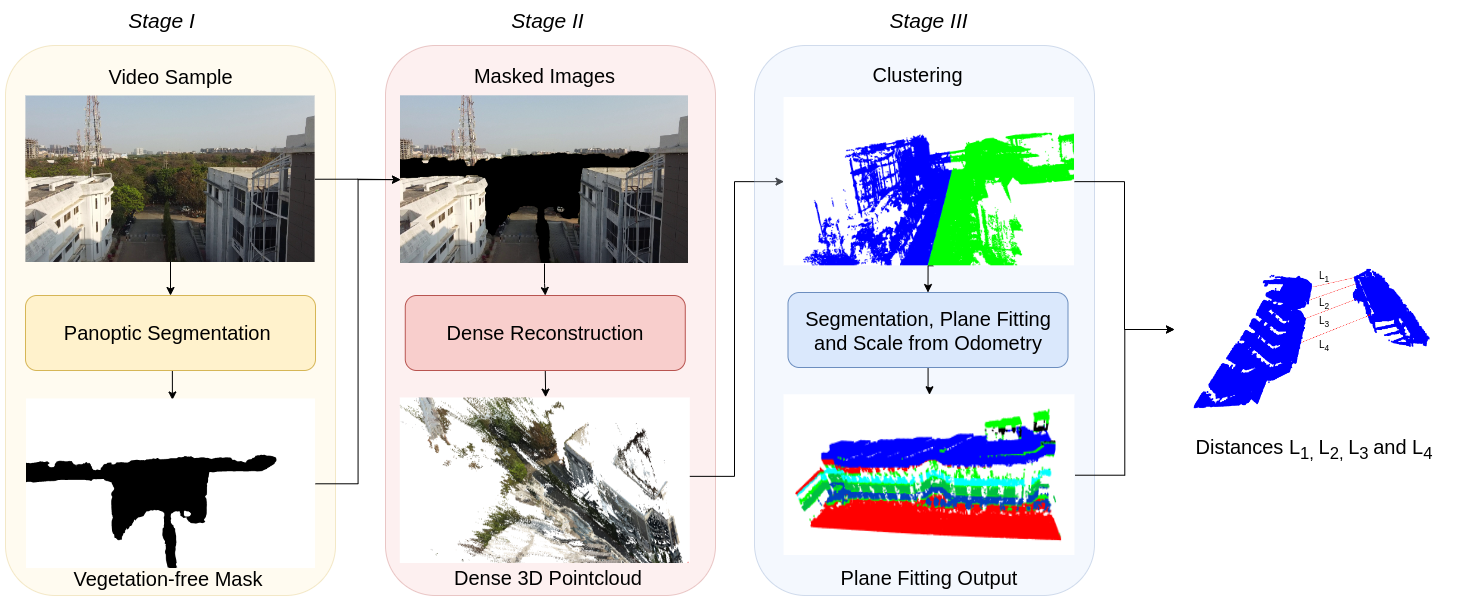}
    \caption{Architecture for estimation of distance between adjacent structures.}
    \label{fig:distance_module_pipeline}
\end{figure}

\begin{figure}
    \subfigure[\textit{Image sample}]
    {
        \includegraphics[width=4cm]{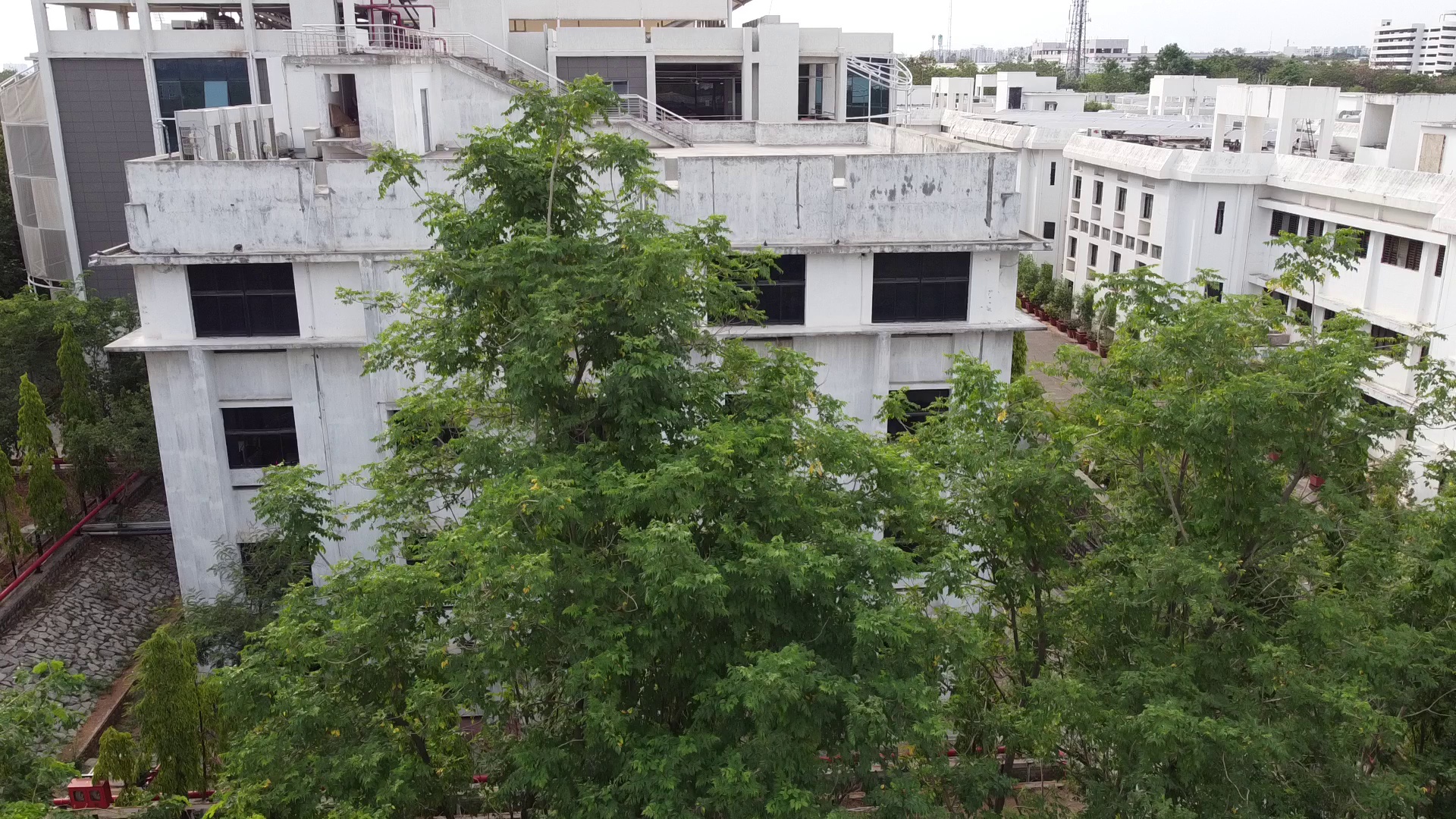}
        \label{fig:panoptic1}
    }
    \subfigure[\textit{Vegetation-free mask}]
    {
        \includegraphics[width=4cm]{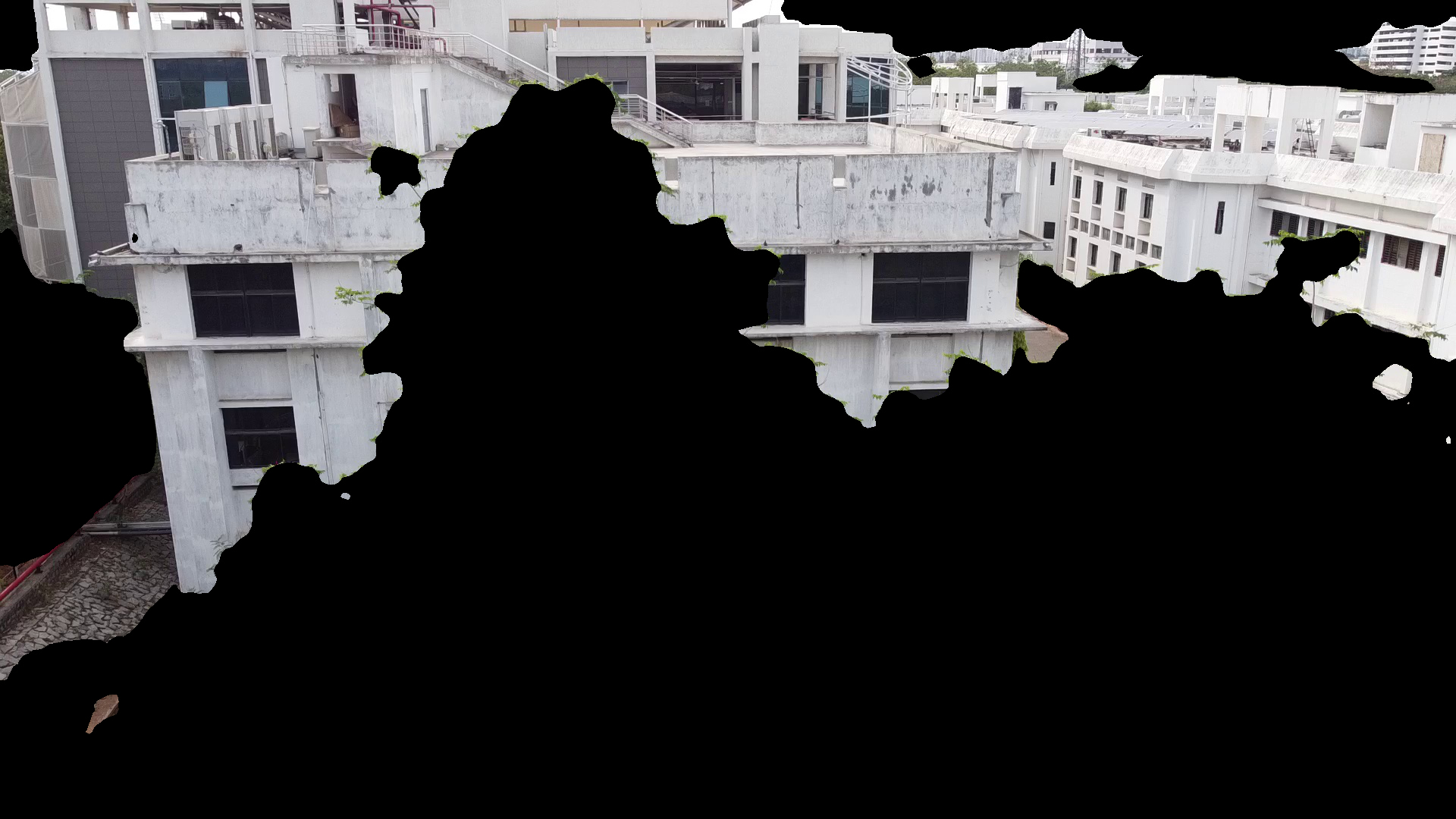}
        \label{fig:panoptic2}
    }
    
    \caption{Removal of vegetation from the sample images enhances the structural features of the buildings in the reconstructed 3D model leading to more accurate results.}
    \label{fig:panopticsegmentation}
\end{figure}

We use plane segmentation to obtain the distance between the two adjacent buildings. We have divided our approach into three stages as presented in Fig. \ref{fig:distance_module_pipeline}. In \textit{Stage I}, images were sampled from the video captured by the UAV and panoptic segmentation was performed using a state-of-the-art network \cite{wu2019detectron2}, to obtain vegetation-free masks. This removes trees and vegetation near the vicinity of the buildings and thus improves the accuracy of our module. Fig. \ref{fig:panopticsegmentation} shows the impact of panoptic segmentation for \textit{frontal mode}. In \textit{Stage II}, the masked images were generated from the binary masks and the corresponding images. The masked images are inputs to a state-of-the-art image-based 3D reconstruction library \cite{schoenberger2016mvs, schoenberger2016mvs1} which outputs a dense 3D point cloud and the camera poses through Structure-from-Motion. Our approach for all the three modes is same for the first two stages.

In \textit{Stage III}, we aim to extract planes from the given point cloud that are essential to identify structures such as roof and walls of the building. We employ the co-ordinate system depicted in Fig. \ref{distance_module}. We can divide this task into two parts: i) Isolation of different building clusters and ii) Finding planes in each cluster. Isolation of the concerned buildings is done using euclidean clustering thereby creating two clusters. For instance, in \textit{Roof Mode} the clusters are distributed on either side of the Y-axis. Similarly, for \textit{In-Between} and \textit{Frontal mode} the clustering happens about the Z-axis. In order to extract the planes of interest, we slice each cluster along a direction parallel to our plane of interest, into small segments of 3D points. For instance, in \textit{Roof mode} we are interested in fitting a plane along the roof of a building; therefore, we slice the building perpendicular to ground normal, i.e, the Z-axis. Finally, Random Sample Consensus (RANSAC) algorithm is applied for each segment of 3D points to iteratively fit a plane and obtain a set of parallel planes as shown under the \textit{Stage III} in Fig. \ref{fig:distance_module_pipeline}.

Our approach selects a plane from the set of planes estimated in \textit{Stage III} for each building based on the highest number of inliers. As stated above, for each mode, the selected planes for the adjacent buildings have the same normal unit vector. Further, we sample points on these planes to calculate the distance between the adjacent buildings at different locations. The scale estimation is done by using the odometry data received from the UAV and the estimated distance is scaled up to obtain the actual distance between the adjacent buildings. This was done by time-syncing the flight logs, that contains GPS, Barometer and IMU readings, with the sampled images.

\subsection{Plan Shape and Roof Area Estimation} \label{sec:planshape}
\begin{figure}
    \centering
    \includegraphics[width=10cm]{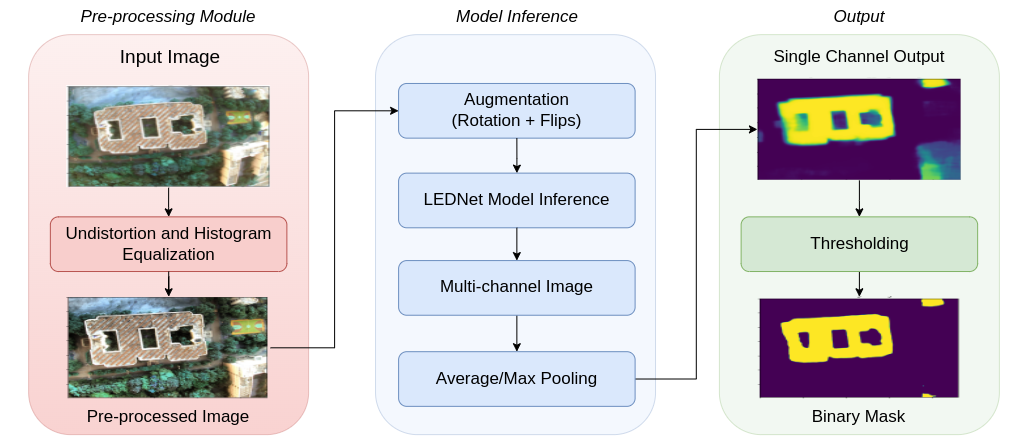}
    \caption{Architecture of the \textit{Plan Shape} module providing the segmented mask of the roof as output from the raw input  image.}
    \label{fig:roofsegmentationlednet}
\end{figure}

The dataset for roof-top of various buildings was collected as described in Section \ref{sec:datacollection}. This dataset was used to estimate the layout and area of the roof through semantic segmentation. The complete \textit{Plan Shape} module has been summarized in Fig. \ref{fig:roofsegmentationlednet}. For the task of roof segmentation, we use a state-of-the-art semantic segmentation model, LEDNet \cite{lednet}. The asymmetrical architecture of this network leads to reduction in network parameters resulting in a faster inference process. The split and shuffle operations in the residual layer enhances information sharing while the decoder's attention mechanism reduces complexity of the whole network. We subject the input images to a pre-processing module that removes distortion from the wide-angle images. Histogram equalization is also performed to improve the contrast of the image. 
Data augmentation techniques (4 rotations of 90\degree + horizontal flip + vertical flip) were used during inference to improve the network's performance and increase robustness. The single-channel grey-scale output is finally thresholded to obtain a binary mask.   The roof area from the segmentation masks can be obtained by using Equation \ref{roofarea} where $C$ is the contour area (in \textit{pixels\textsuperscript{2}}), obtained from the segmented mask, $D$ is the depth of the roof from the camera (in \textit{m}) and $f$ is the focal length of the camera (in \textit{pixels}) used.
\begin{equation}
    Area \,(m^2) = C \times (D/f)^2 
    \label{roofarea}
\end{equation}

\subsection{Roof Layout Estimation}
\begin{figure}
    \centering
    \includegraphics[height=5cm, width=10cm]{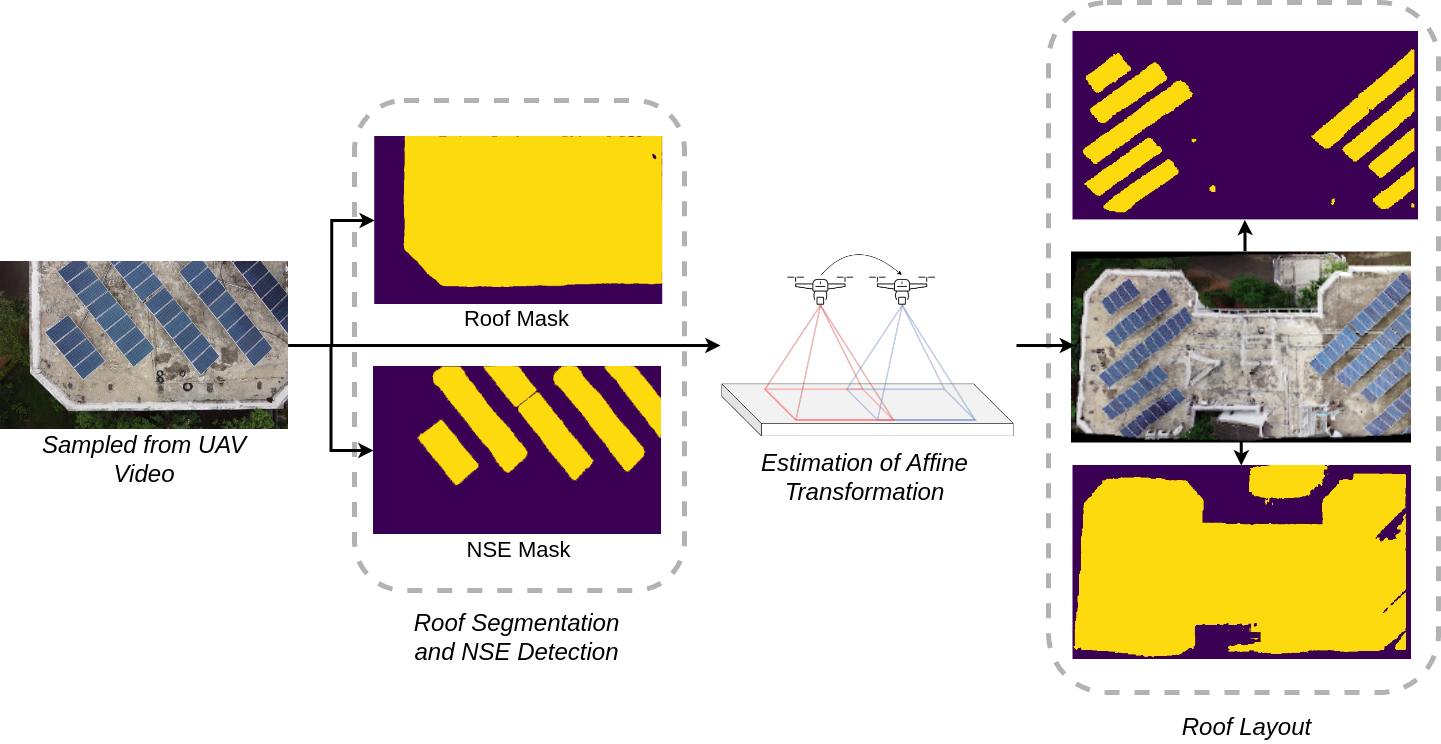}
    \caption{We exploit the planarity of the roof and the fact that the distance between the UAV and the roof will be constant (the UAV is flown at a constant height). This enables us to relate two consecutive images through an affine transformation.}
    \label{fig:planediagram}
\end{figure}

The data for this module was collected as described in Section \ref{sec:datacollection}. Due to the camera FOV limitations and to maintain good resolution, it is not possible to capture the complete view of the roof in a single image, especially in the case of large sized buildings. Hence, we perform large scale stitching of partially visible roofs followed by NSE detection and roof segmentation. Fig \ref{fig:planediagram} shows the approach adopted for \textit{Roof Layout Estimation}.

\subsubsection{Large Scale Aerial Image Stitching:}
 We exploit the planarity of the roof and the fact that the UAV is flown at a constant height from the roof. Instead of opting for homography, that relates two geometric views in case of image stitching, we opt for affine transformations. Affine transformations are linear mapping methods that preserve points, straight lines, and planes.

Let $I = \{i_1, i_2, i_3, ... , i_N\}$ represent an ordered set of images sampled from a video collected as per Section \ref{sec:datacollection}. The image stitching algorithm implemented has been summarized below:
\begin{enumerate}
    \item Features were extracted in image $i_1$, using ORB feature detector and tracked in the next image, $i_{2}$ using optical flow. This helped in effective rejection of outliers.
    \item The obtained set of feature matches across both the images were used to determine the affine transformation matrix using RANSAC.
    \item Images $i_{1}$ and $i_{2}$ were then warped as per the transformation and stitched on a \textit{canvas}. 
    \item Affine transformation was calculated between image $i_3$ and the previously warped image $i_2$ before it was stitched using steps 1 and 2. Image $i_3$ was then warped and stitched on the same \textit{canvas}. 
    \item Step 4 was repeated for the next set of images, that is, affine transformation was calculated for image $i_4$ and the previously warped image $i_3$ before it was stitched.
\end{enumerate}

\subsubsection{Detecting Objects on Rooftop:}
For identification of NSE on the rooftop, we use a state-of-the art object detection model, Detic \cite{zhou2021detecting} because it is highly flexible and has been trained for large number of classes. In order to estimate the roof layout, it is essential to detect and locate the NSE as well as the roof from a query image. Note that we classify all the NSE as a single class. This information can then be represented as a semantic mask which will be to calculate the percentage of occupancy of the NSE. A custom vocabulary comprising of the NSE was passed to the model. The roof was segmented out using LEDNet as described in Section \ref{sec:planshape}.


\section{Results}
This section presents the results for the different modules of automated building inspection using aerial images.
\subsection{Distance Between Adjacent Buildings}
\begin{figure}
    \subfigure[\textit{Roof mode}]
    {
        \includegraphics[width=3.7cm]{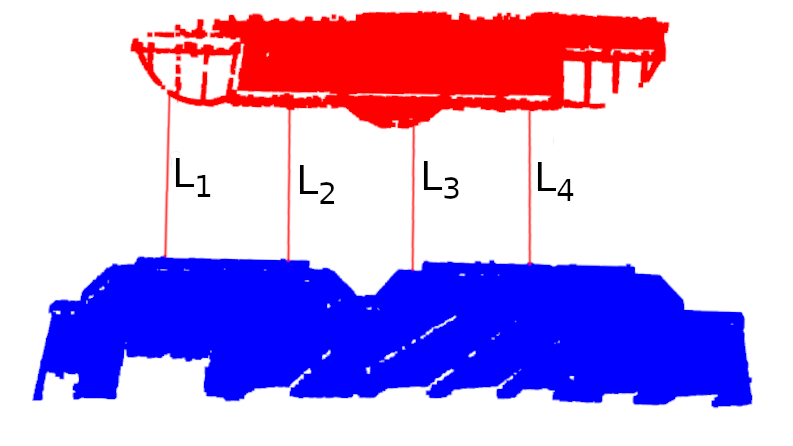}
        \label{fig:res1}
    }
    \subfigure[\textit{In-Between mode}]
    {
        \includegraphics[width=3.7cm]{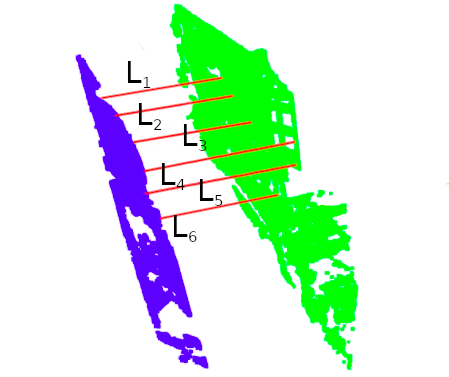}
        \label{fig:res2}
    }
    \subfigure[\textit{Frontal mode}]
    {
        \includegraphics[width=3.7cm]{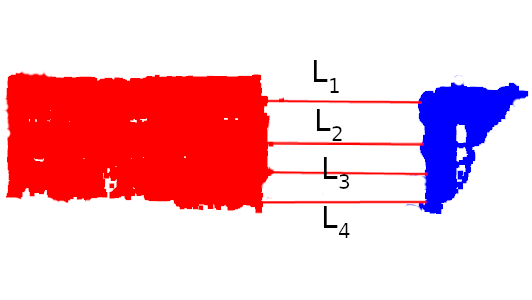}
        \label{fig:res3}
    }\\
    \subfigure[\textit{Roof mode}]
    {
        \includegraphics[width=3.7cm]{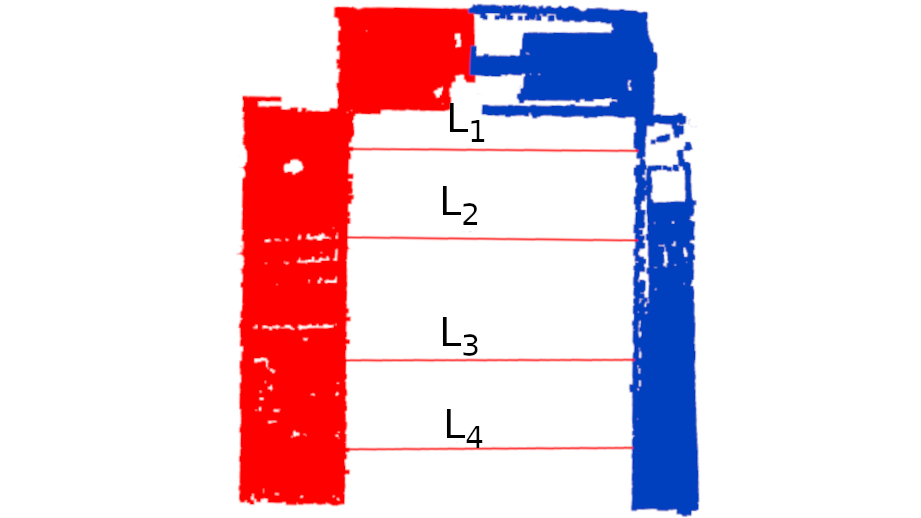}
        \label{fig:res4}
    }
    \subfigure[\textit{In-Between mode}]
    {
        \includegraphics[width=3.7cm]{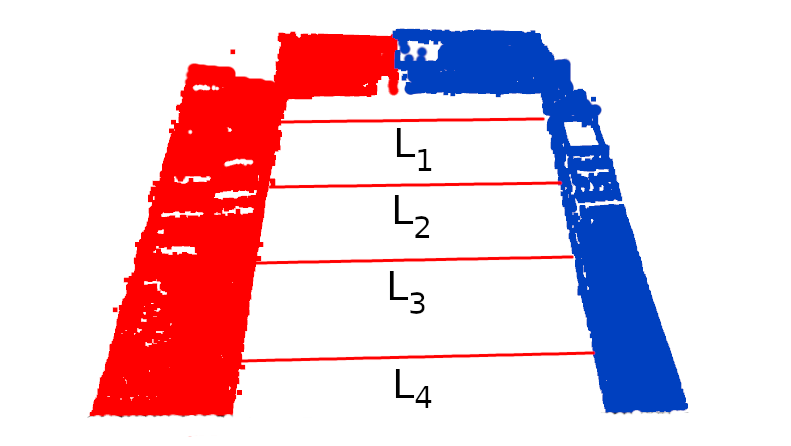}
        \label{fig:res5}
    }
    \subfigure[\textit{Frontal mode}]
    {
        \includegraphics[width=3.7cm]{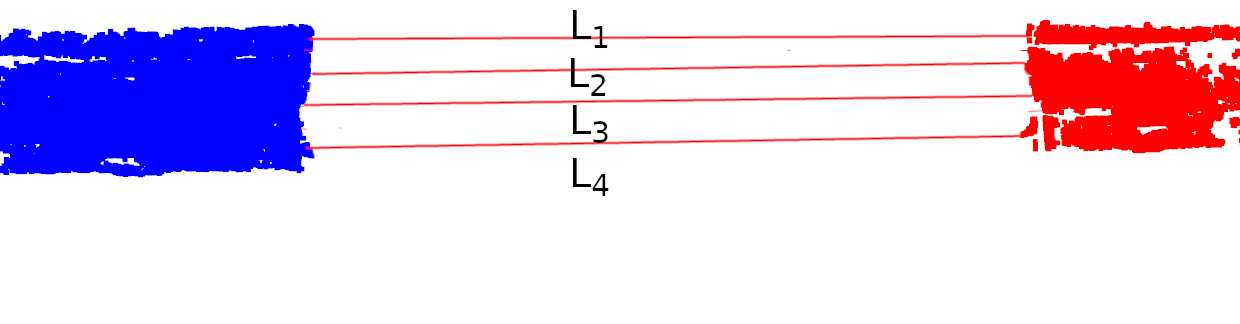}
        \label{fig:res6}
    }

    \caption{\ref{fig:res1}, \ref{fig:res2} and \ref{fig:res3} and \ref{fig:res4}, \ref{fig:res5} and \ref{fig:res6} represent the implementation of plane fitting using piecewise-RANSAC in different views for \textit{Buildings 1, 2} and \textit{Building 3} respectively.}
    \label{fig:distanceresults}
\end{figure}

We validated our algorithm on real aerial datasets of adjacent buildings and structures. In particular, we tested all the modes of this module on a set of adjacent buildings, \textit{Buildings 1 and 2}, and also on \textit{Building 3}, a U-shaped building. The resulting distances for all the modes can be visualized through Fig. \ref{fig:distanceresults}. The corresponding distances visualized in Fig. \ref{fig:distanceresults} have been documented in Table \ref{table:distanceresultsbuilding12} and \ref{table:distanceresultsbuilding3}. We obtain the ground truth from using a Time-of-Flight (ToF) based range measuring sensor\footnote{The ToF sensor can be found at: \href{https://www.terabee.com/shop/lidar-tof-range-finders/teraranger-evo-60m/}{https://www.terabee.com/shop/lidar-tof-range-finders/teraranger-evo-60m/}}. This sensor has a maximum range of 60 meters. We also compare the results with that from \textit{Google Earth}. It must be noted that using \textit{Google Earth}, it is not possible to measure some distances due to lack of 3D imagery.


\setlength{\tabcolsep}{1pt}
\newcolumntype{P}[1]{>{\centering\arraybackslash}p{#1}}
\begin{table}
\caption{Distances calculated for \textit{Building 1 and 2} using our method and Google Earth for all three modes.}
\label{table:distanceresultsbuilding12}
\centering
  \begin{tabular}{|P{1.6cm}|P{2.2cm}|P{1.5cm}|P{1.5cm}|P{1.5cm}|P{1.5cm}|P{1.5cm}|}

    \hline
    \multirow{2}{*}{Mode} &
      \multicolumn{6}{c|}{\textbf{\textit{Buildings 1 and 2}}} \\
      \cline{2-7}
   & Distance Reference & Ground Truth & Google Earth & Estimated & Error (Google Earth) & Error (Estimated)\\
    \hline
    \multirow{4}{*}{\textit{Roof}}& L\textsubscript{1} in Fig \ref{fig:res1}& 16.40 m & 17.14 m & 16.70 m & 4.5\% & \textbf{1.8\%}  \\
    \cline{2-7}
     & L\textsubscript{2} in Fig \ref{fig:res1}  & 12.96 m & 12.91 m & 12.94 m & 0.3\% & \textbf{0.15\%} \\
     \cline{2-7}
     & L\textsubscript{3} in Fig \ref{fig:res1} & 12.01 m & 12.08 m & 11.97 m & 0.58\% & \textbf{0.33\%}  \\
     \cline{2-7}
     & L\textsubscript{4} in Fig \ref{fig:res1} & 13.30 m & 13.00 m & 12.77 m & \textbf{2.31\%} & 3.98\% \\
     \hline
     
      \multirow{5}{*}{\textit{In-Between}}& L\textsubscript{1} in Fig \ref{fig:res2} & 13.31 m & 13.50 m & 13.22 m & 1.42\% & \textbf{0.67\%}  \\
      \cline{2-7}
      & L\textsubscript{2} in Fig \ref{fig:res2} & 12.91 m & 12.40 m & 12.87 m & 3.95\% & \textbf{0.31}\%   \\
    \cline{2-7}
     & L\textsubscript{3} in Fig \ref{fig:res2} & 12.30 m & 12.43 m & 12.12 m & \textbf{1.00\%} & 1.39\% \\
     \cline{2-7}
    & L\textsubscript{4} in Fig \ref{fig:res2} & 12.70 m & 12.87 m & 12.50 m & \textbf{1.33\%} & 1.57\%   \\
     \cline{2-7}
     & L\textsubscript{5} in Fig \ref{fig:res2} & 13.95 m & 13.84 m & 13.87 m & 0.79\% & \textbf{0.51\%}   \\
     \cline{2-7}
     & L\textsubscript{6} in Fig \ref{fig:res2} & 12.60 m & 12.69 m & 12.56 m &  0.71\% &  \textbf{0.31\%}   \\
     \hline
      \multirow{4}{*}{\textit{Frontal}} & L\textsubscript{1} in Fig \ref{fig:res3} & 16.96 m & 16.91 m & 16.92 m & 0.29\% & \textbf{0.23\%} \\
    \cline{2-7}
     & L\textsubscript{2} in Fig \ref{fig:res3} & 16.96 m & -  & 16.78 m & - & \textbf{1.06\%} \\
     \cline{2-7}
     & L\textsubscript{3} in Fig \ref{fig:res3} & 16.96 m & - & 17.13 m & - & \textbf{1.00\%}\\
     \cline{2-7}
     & L\textsubscript{4} in Fig \ref{fig:res3} & 16.96 m & -  & 17.05 m & - & \textbf{0.53\%} \\
     \hline
  \end{tabular}
\end{table}

\setlength{\tabcolsep}{1pt}
\begin{table}

\caption{Distances calculated for \textit{Building 3} using our method and Google Earth for all three modes. }
\label{table:distanceresultsbuilding3}
\centering
  \begin{tabular}{|P{1.6cm}|P{2.2cm}|P{1.5cm}|P{1.5cm}|P{1.5cm}|P{1.5cm}|P{1.5cm}|}

    \hline
    \multirow{2}{*}{Mode} &
      \multicolumn{6}{c|}{\textbf{\textit{Building 3}}} \\
      \cline{2-7}
   & Distance Reference & Ground Truth & Google Earth & Estimated & Error (Google Earth) & Error (Estimated)\\
    \hline
    \multirow{4}{*}{\textit{Roof}}& L\textsubscript{1} in Fig \ref{fig:res4} & 33.28 m & 33.58 m & 33.26 m & 0.90 \% & \textbf{0.06 \%}  \\
    \cline{2-7}
     & L\textsubscript{2} in Fig \ref{fig:res4} & 33.28 m & 33.63 m & 33.22 m & 1.05 \% & \textbf{0.18 \%} \\
     \cline{2-7}
     & L\textsubscript{3} in Fig \ref{fig:res4} & 33.28 m & 33.00 m & 33.28 m & 0.84 \% & \textbf{0.00 \%}  \\
     \cline{2-7}
     & L\textsubscript{4} in Fig \ref{fig:res4} & 33.28 m & 33.35 m & 33.81 m & \textbf{0.21 \%} & 1.59 \% \\
     \hline
     
      \multirow{3}{*}{\textit{In-Between}}& L\textsubscript{1} in Fig \ref{fig:res5} & 33.28 m & 32.58 m & 33.11 m & 2.10 \% & \textbf{0.51 \%}  \\
      \cline{2-7}
      & L\textsubscript{2} in Fig \ref{fig:res5} & 33.28 m & 33.12 m & 32.40 m & \textbf{0.48 \%} & 2.64 \%   \\
    \cline{2-7}
     & L\textsubscript{3} in Fig \ref{fig:res5} & 33.28 m & 32.94 m & 32.78 m & \textbf{1.02 \%} & 1.50 \%  \\
     \cline{2-7}
     & L\textsubscript{4} in Fig \ref{fig:res5} & 33.28 m & 32.25 m & 32.81 m & 3.09 \% & \textbf{1.41 \%}   \\
     \hline
      \multirow{4}{*}{\textit{Frontal}} & L\textsubscript{2} in Fig \ref{fig:res6} & 33.28 m & 33.57 m  & 33.26 m & 0.87 \% & \textbf{0.06 \%} \\
    \cline{2-7}
     & L\textsubscript{1} in Fig \ref{fig:res6} & 33.28 m & -  & 33.60 m & -  & \textbf{0.96 \%} \\
     \cline{2-7}
     & L\textsubscript{3} in Fig \ref{fig:res6} & 33.28 m & - & 33.59 m & - & \textbf{0.93} \%\\
     \cline{2-7}
     & L\textsubscript{4} in Fig \ref{fig:res6} & 33.28 m & -  & 33.99 m & - & \textbf{2.13} \% \\
     \hline
  \end{tabular}

\end{table}
\setlength{\tabcolsep}{1.4pt}

\subsection{Plan Shape and Roof Area Estimation}
\begin{figure}
    \centering
    \includegraphics[width=12cm]{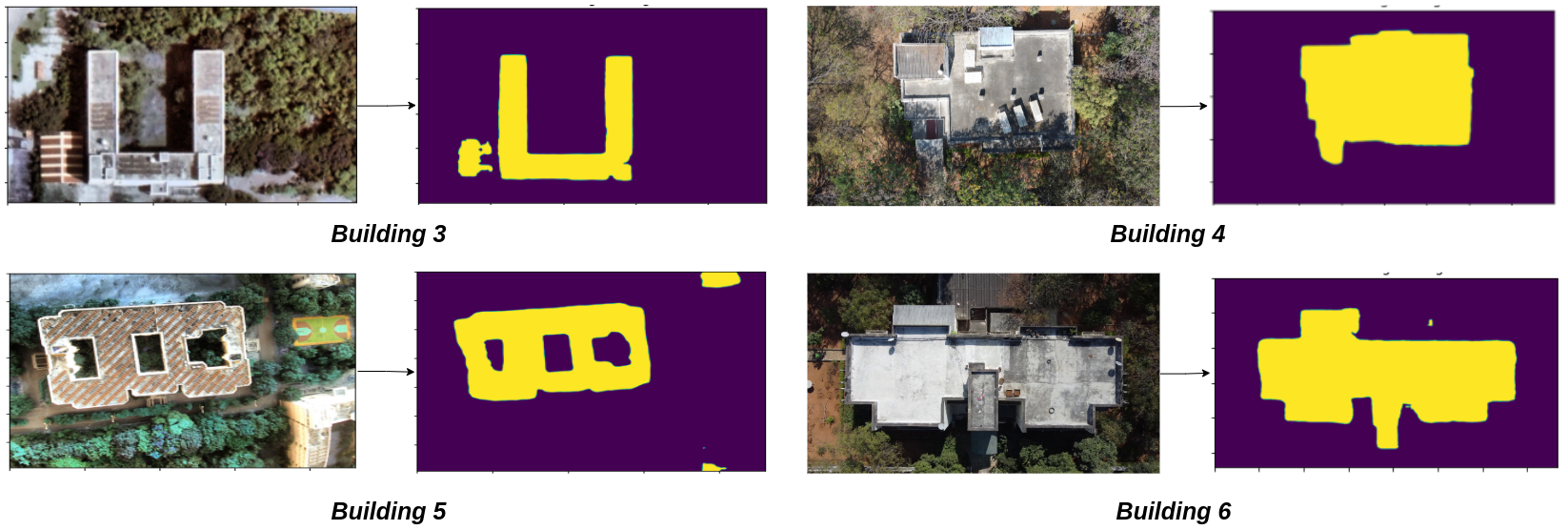}
    \caption{Roof Segmentation results for 4  buildings.}
    \label{fig:RoofSegmentationResults}
\end{figure}

The roof area was estimated from images taken at different depths, that is, when the UAV was operated at different altitudes ranging from 50m to 100m. The module was tested on various campus buildings. The results in Table \ref{table:areaestimation} were averaged out for all samples corresponding to the same building. The module estimates the roof area with an average difference of 4.7\% with Google Earth data. Predicted roof masks of some buildings from LEDNet are shown in Fig \ref{fig:RoofSegmentationResults}.
\begin{figure}
    \centering
    \includegraphics[width=12cm]{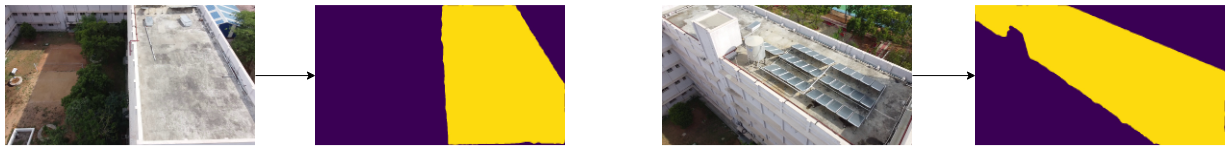}
    \caption{We use LEDNet in both \textit{Plan Shape and Roof Area Estimation} as well as \textit{Roof Layout Estimation.} The trained model correctly segments out the roof in case of a non-orthogonal view, that is, when the downward-facing camera is not directly above the roof.}
    \label{fig:NonOrthogonalesults}
\end{figure}

\setlength{\tabcolsep}{1.4pt}
\begin{table}
\begin{center}
\caption{Roof Area Estimation Results}
\label{table:areaestimation}
\begin{tabular}{|P{2.3cm}|P{2.3cm}|P{2.3cm}|P{2.3cm}|P{2.3cm}|}
\hline
\textbf{Building} & \textbf{Area Measured using Google Earth} & \textbf{Estimated Area} & \textbf{Absolute Difference} & \textbf{Percentage Difference}\\
\hline
\textit{Building 3}& 1859.77 m\textsuperscript{2} & 1939.84 m\textsuperscript{2} & 80.07 m\textsuperscript{2} & 4.3 \% \\
\hline
\textit{Building 4}& 350 m\textsuperscript{2}   & 331.30 m\textsuperscript{2} & 18.70 m\textsuperscript{2} & 5.3 \%\\
\hline
\textit{Building 5}& 340 m\textsuperscript{2} & 329.55 m\textsuperscript{2} & 10.45 m\textsuperscript{2} & 3.1 \%\\
\hline
\textit{Building 6}& 3,127.60 m\textsuperscript{2} & 2936.82 m\textsuperscript{2} & 190.78 m\textsuperscript{2} & 6.1 \%\\
\hline
\end{tabular}
\end{center}
\end{table}

\subsection{Roof Layout Estimation}
\begin{figure}
    \centering
    \subfigure[Stitched Image]
    {
    \includegraphics[height=1.35cm,width=11.5cm]{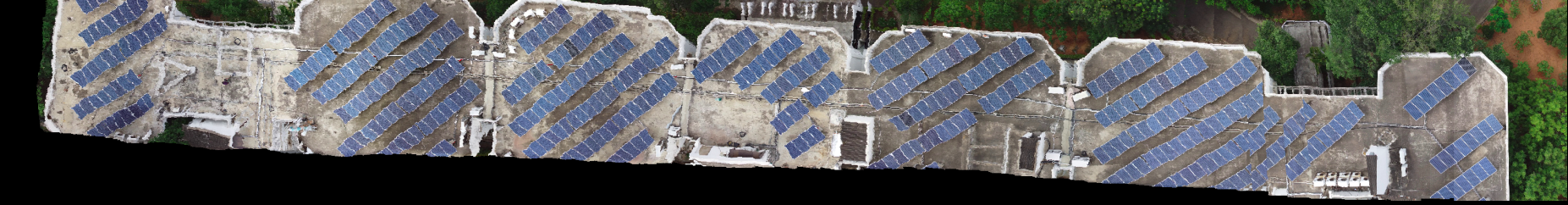}

    \label{fig:imagestitchingresults}}
    \subfigure[Roof Mask]
    {
    \includegraphics[height=1.35cm,width=11.5cm]{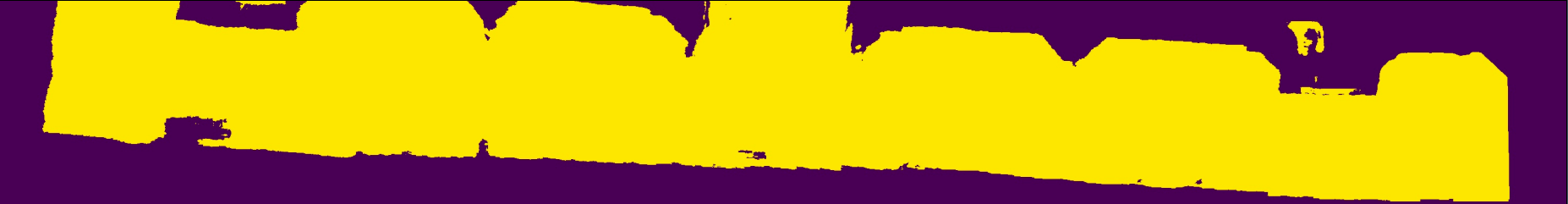}

    \label{fig:roofmaskresults}}
    \subfigure[Object Mask]
    {
    \includegraphics[height=1.35cm,width=11.5cm]{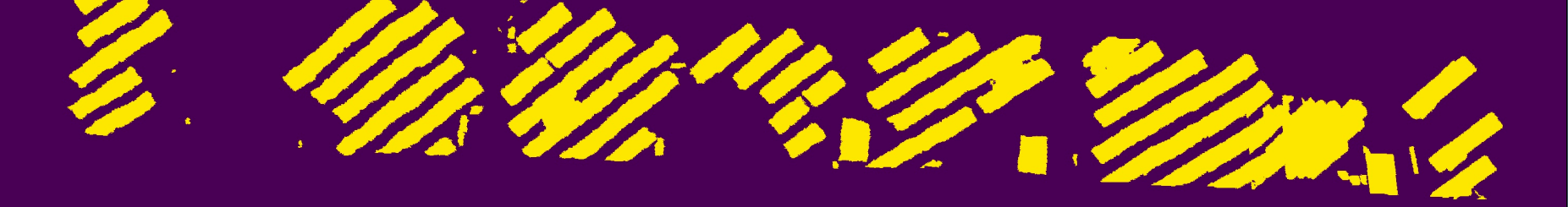}
    
    \label{fig:objectmaskresults}}
    \caption{Results for Roof Layout Estimation.}
\end{figure}

Data for \textit{Roof Layout Estimation} was collected as described in Section \ref{sec:datacollection}. Images were sampled at a frequency of $10Hz$.  The video collected for \textit{Roof Layout Estimation} was sampled at a frequency of $1\,Hz$ generating 98 images. The results of image stitching can be visualized in Fig. \ref{fig:imagestitchingresults} with the corresponding roof mask in Fig. \ref{fig:roofmaskresults} and the NSE mask in Fig. \ref{fig:objectmaskresults}. The percentage occupancy was calculated by taking the ratio of object occupancy area (\textit{pixels\textsuperscript{2}}) in Fig. \ref{fig:objectmaskresults} to total roof area (\textit{pixels\textsuperscript{2}}) in Fig. \ref{fig:roofmaskresults}. The final percentage occupancy obtained was $38.73\%$.


\section{Discussion}
We estimate the distance between adjacent structures using 3D reconstruction and conditional plane fitting and validate its performance on ground truth data from a ToF sensor. We also make a comparison of our proposed module with Google Earth and validate our superior performance. Moreover, it is not possible to employ Google Earth for this module universally due to the lack of 3D imagery for all the buildings. We validated our distance estimation algorithm and compared the results with the ground truth and Google Earth. We estimated the distance between adjacent structures with an average error of 0.94\%, which is superior to Google Earth which performs with an average error of 1.36\%.  Our rooftop area estimation module performs with an average difference of 4.7\% when compared to Google Earth. It is observed that the difference remains near-constant irrespective of the size of the rooftop area. Considering the irregular shape of the roofs, it is challenging to measure the ground truth of the roof area manually and it is also error-prone and resource-intensive. The roof layout was estimated through semantic segmentation, object detection, and large-scale image stitching of 98 images. We also detected NSE on the rooftops and found their percentage occupancy to be 38.73\%.
  

\section{Conclusion} 
This paper presented an implementation of a considerable amount of approaches that have been developed, aiming at modeling the structure of buildings. Seismic risk assessment of buildings involves the estimation of several structural parameters. It is important to estimate the parameters which can describe the geometry of buildings, plan shape of the rooftops, and size of the buildings. In particular, we estimated the distances between adjacent buildings and structures, plan shape of a building, roof area, and percentage of area occupied by NSE. We plan to release these modules in the form of an open-source library that can be easily used by non-computer vision experts. Future work includes quantifying the flatness of ground, crack detection, and identification of water tanks and staircase exits that could help in taking preliminary precautions for earthquakes.


\paragraph{\textbf{Acknowledgement:}}The authors acknowledge the financial support provided by IHUB, IIIT Hyderabad to carry out this research work under the project: IIIT-H/IHub/Project/Mobility/2021-22/M2-003.
%
%
\bibliographystyle{unsrtnat}
\bibliography{egbib}
\end{document}